\documentclass[pmlr]{jmlr}

\usepackage{appendix}
\usepackage{booktabs}
\RequirePackage{graphicx}
\usepackage{tabularx}
\usepackage[labelfont=bf]{caption} 
 \usepackage{booktabs}
\usepackage{longtable}
\usepackage{makecell}

\makeatletter
\def\set@curr@file#1{\def\@curr@file{#1}} 
\makeatother
\usepackage[load-configurations=version-1]{siunitx} 

\title[Zero-Shot Clinical Trial Matching with LLMs]{Zero-Shot Clinical Trial Patient Matching with LLMs}

\author{\Name{Michael Wornow*} \Email{mwornow@stanford.edu}\\
 \Name{Alejandro Lozano*} \Email{lozanoe@stanford.edu}\\
 \Name{Dev Dash} \Email{dev@stanford.edu}\\
 \Name{Jenelle Jindal} \Email{jjindal@stanford.edu}\\
 \Name{Kenneth W. Mahaffey} \Email{kmahaf@stanford.edu}\\
 \Name{Nigam H. Shah} \Email{nshah@stanford.edu}\\
 \addr Stanford University, Palo Alto, CA, USA
}

\begin{document}

\maketitle

\begin{abstract}
\noindent \textbf{Background:} Matching patients to clinical trials is a key unsolved challenge in bringing new drugs to market. Today, identifying patients who meet a trial's eligibility criteria is highly manual, taking up to 1 hour per patient. Automated screening is challenging, however, as it requires understanding unstructured clinical text. \\ 
\noindent \textbf{Methods:} We design a zero-shot LLM-based system which, given a patient's medical history as unstructured clinical text, evaluates whether that patient meets a set of trial inclusion criteria (also specified as free text). We investigate different prompting strategies and design a novel two-stage retrieval pipeline to reduce the number of tokens processed by up to a third while retaining high performance. \\
\noindent \textbf{Results:} We achieve state-of-the-art performance on the n2c2 2018 cohort selection challenge, the largest clinical trial patient matching public benchmark. Second, we show that our system can improve the data and cost efficiency of matching patients an order of magnitude faster and more cheaply than the status quo. Third, we measure the interpretability of our system by having clinicians evaluate the natural language justifications generated for each eligibility decision, and show that it can output coherent explanations for 97\% of its correct decisions and even 75\% of its incorrect ones. \\
\noindent \textbf{Conclusion:} Our results establish the feasibility of using LLMs to accelerate clinical trial operations. Our zero-shot retrieval architecture enables scaling to arbitrary trials and patient record length with minimal reconfiguration.

\end{abstract}

\section{Introduction}
\label{sec:intro}

\begin{figure*}[!ht]
\includegraphics[width=1.0\textwidth]{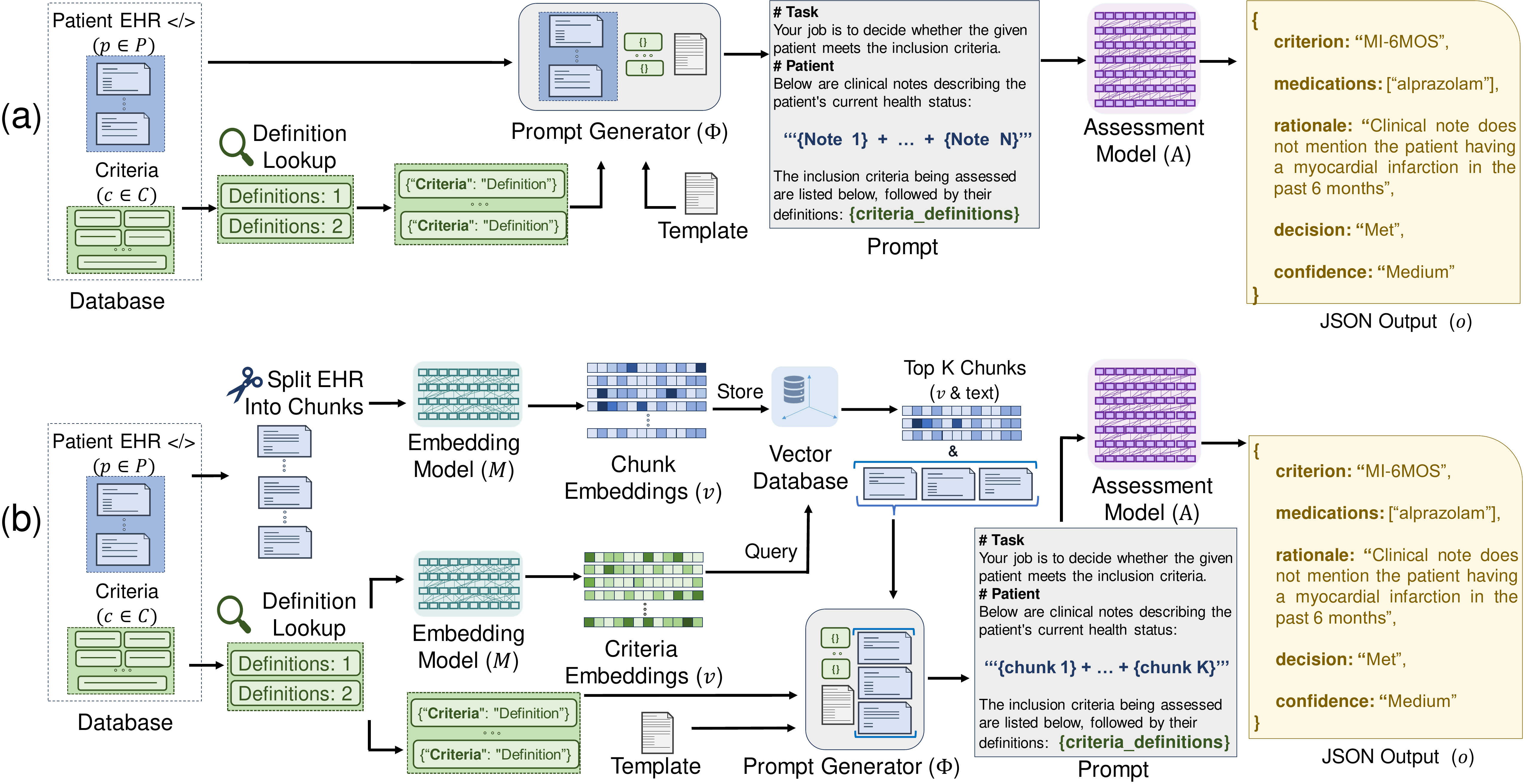}
\caption{\textbf{Zero-Shot system design}}
\medskip
\small
We explore zero-shot clinical trial patient matching with LLMs under two system designs:
(a) We inject the patient's notes into a prompt that is fed to an Assessment LLM (e.g. GPT-4) for evaluation. The \textit{ACAN} prompting strategy is depicted here. (b) In our two-stage retrieval pipeline, we first query the top-$k$ most relevant chunks from the patient's notes, then inject only those top-$k$ chunks into the prompt fed to the Assessment LLM. Again, the \textit{ACAN} prompting strategy is depicted here.
\label{fig:figure_1}
\end{figure*}

Identifying eligible patients for clinical trials is a key bottleneck in advancing new drugs to market \citep{woo2019ai}. One third of clinical trials fail because they cannot enroll enough patients, and recruitment costs an average of 32\% of a trial’s budget \citep{Deloitte}. 

For patients, enrolling in a trial can bring several benefits, such as access to novel therapies, increased monitoring from expert care teams, and better overall health outcomes \citep{ctbenefits, bouzalmate2022benefits}. Unfortunately, 94\% of patients are never informed by their doctors about trials for which they might qualify \citep{nuttall2012considerations}.

That is because identifying patients who are eligible for a trial is often highly manual and time-consuming \citep{brogger2020online}. Each trial has a long list of \textit{eligibility criteria} which must be exactly met by every patient who enrolls \citep{us2020enhancing}. To identify these patients, a trained \textit{clinical research coordinator} will manually review hundreds of patients' electronic health records (EHRs) \citep{penberthy2012effort, calaprice2020improving}. Over 80\% of the relevant information is stored as unstructured text -- e.g. progress notes, emails, radiology reports, and genetic tests \citep{kong2019managing} -- making it difficult to automatically process \citep{leaman2015challenges}. As a result, eligibility screening for a Phase III cancer trial can take almost an hour per patient \citep{penberthy2012effort}.

Traditional natural language processing (NLP) approaches have seen limited success with clinical text due to its idiosyncratic grammar and terminology \citep{leaman2015challenges}. Large language models (LLMs) represent a key inflection point in NLP capabilities \citep{brown2020language}, and thus offer a promising approach for accelerating patient recruitment. 

In this paper, we aim to build upon recent work exploring the application of LLMs to patient matching \citep{trialgpt, wong2023scaling, trialllama}. In contrast to this prior work, our primary focus is on the \textit{efficiency} of LLM-based approaches -- namely, their cost in terms of data, money, and time. These considerations are essential if such systems aim to be deployed at scale given the limited resources of most health systems.

Our contributions are as follows.

\begin{enumerate}
    \item We investigate the \textbf{zero-shot} (i.e. out-of-the-box) performance of LLMs on evaluating patient eligibility. The zero-shot approach that we study offers several advantages over alternative such as fine-tuning or few-shot prompting, as the former requires no labeled data and would thus allow our system to more readily scale to arbitrary trials. It is also significantly cheaper than fine-tuning (which require purchasing compute) or few-shot prompting (which inflates token counts by a factor of over 2x in experiments). Our zero-shot GPT-4-based system \textbf{achieves state-of-the performance} on Track 1 of the 2018 n2c2 cohort selection benchmark, which is the largest and most realistic publicly available clinical trial patient matching benchmark \citep{stubbs2019cohort}.
    \item We show how to improve the \textbf{data and cost efficiency} of LLM-based systems. We identify prompting strategies that best balance performance and cost. We also design a \textbf{two-stage retrieval-based system}, as shown in Figure \ref{fig:figure_1}, which pre-filters relevant clinical notes and can reduce token usage by over a third while still beating the state-of-the-art on certain metrics.
    \item We demonstrate the potential for \textbf{human-in-the-loop deployment} by having clinicians evaluate the natural language justifications generated by the LLM for each assessment it makes. We find that GPT-4 is able to output a coherent rationale for its correct decisions 97\% of the time, and even when it makes an incorrect decision, our annotators agreed with 75\% of its justifications.
\end{enumerate}

The rest of the paper is structured as follows. First, we provide an overview of related work. Second, we detail the models that we test, our different prompting strategies, and our retrieval system. Third, we describe our dataset and evaluation set-up. Fourth, we report our system's results. Fifth, we discuss limitations and propose areas for further research.

\section{Related Work}

\subsection{Large Language Models (LLMs)}

Natural language processing (NLP) was revolutionized by the advent of deep learning models which could learn from huge amounts of unlabeled text, e.g. all of Wikipedia \citep{howard2018universal}. As these models increased in size and their datasets became even larger (e.g. all of the Internet), their capabilities continued to improve \citep{brown2020language, kaplan2020scaling}. Known as ``large language models" (LLMs), they have surpassed humans on a number of benchmarks \citep{team2023gemini} and can now conduct diagnostic conversations with patients \citep{tu2024towards}, translate clinical jargon into layman's terms \citep{mirza2024using}, and perform chart review \citep{goel2023llms}. Most impressively, LLMs have tackled many of these tasks without ever explicitly being trained to solve them \citep{brown2020language}. 

Asking an LLM to solve a task without providing any training examples is referred to as ``zero-shot" learning \citep{xian2018zero}. This stands in sharp contrast to traditional machine learning models which rely on labeled training datasets, as well as conventional rule-based systems which require domain experts to hand-craft programs. In contrast, LLMs such as GPT-4 \citep{gpt4} can be adapted for a wide range of tasks expressed via text. This makes them promising for the task of patient matching.

\subsection{Clinical Trial Patient Matching}

Initial work in automated trial matching focused on automatically transforming free text eligibility criteria into structured SQL queries, which could then be run over the structured data within patient EHRs. Systems such as \textbf{EliXR} \citep{weng_elixr_2011} accomplished this via rule-based algorithms, while \textbf{EliIE} \citep{kang_eliie_2017} and \textbf{Criteria2Query} \citep{yuan_criteria2query_2019} used more advanced NLP methods such as named entity recognition, relation extraction, and negation detection. Most relevant to this paper, the \textbf{RBC} model from \citep{sota} achieved the highest score on the 2018 n2c2 cohort selection benchmark. While well-suited for the specific trials for which they are developed, these systems are difficult to maintain and tend not to generalize  \citep{gehrmann2017comparing}.

More recently, focus has shifted to developing end-to-end systems that can directly match patients to free text criteria. Works such as \textbf{COMPOSE} \citep{gao_compose_2020} and \textbf{DeepEnroll} \citep{zhang_deepenroll_2020} used neural networks to jointly encode enrollment criteria and patient EHRs. Both of these models, however, only incorporated structured EHR data and ignored clinical text. 

\textbf{TrialGPT} was the first end-to-end LLM-based system for processing both unstructured clinical text and eligibility criteria \citep{trialgpt}, and achieved high accuracy. However, they did not conduct zero-shot evaluation and only tested GPT-3.5, which as we later demonstrate missed out on large performance gains. \textbf{Scaling with LLMs} \citep{wong2023scaling} used an GPT-4 to convert free text criteria into structured queries, but did not directly use the LLM to evaluate patients. More broadly, \textbf{TRIALSCOPE} offered an exciting vision of \textit{simulating} entire trials with real world evidence, but did not develop their own patient matching solution \citep{gonzalez2023trialscope}. \textbf{TrialLlama} \citep{trialllama} fine-tuned the open source Llama2 model \citep{touvron2023llama} for trial matching, but this required extensive finetuning using 16 A100s and did not consider zero-shot settings. Additionally, none of these works considered the use of a preliminary retrieval step to first curate the notes fed into the LLM. 

\section{Methods}

In the following, assume we are given a set of patients $\mathcal{P}$ and notes $\mathcal{N}$, where each patient $p \in \mathcal{P}$ has a set of notes $\mathcal{N}_p \subseteq \mathcal{N}$. We have a set $\mathcal{C}$ of trial criteria. There is some unknown function $E: \mathcal{C} \times \mathcal{P} \rightarrow \{0, 1\}$ that maps each criterion and patient to a binary indicator of whether that patient meets (1) or does not meet (0) that criterion. Our goal is to calculate $E(c, p)$ for all patients $p$ and criteria $c$.

Assume we have a small ``embedding" model $\mathbf{M}: \mathcal{T} \rightarrow \mathcal{V}$ which maps some text $t \in \mathcal{T}$ to a vector $v \in V$ and can be run quickly at virtually no cost. We also have a larger downstream ``assessment" model $\mathbf{A}: \mathcal{T} \rightarrow \mathcal{O}$ which maps a prompt $t \in \mathcal{T}$ to some output $o \in \mathcal{O}$ and is expensive and slow. Finally, we have some function $\phi$ which takes a variable number of criteria and patient notes (depending on the prompting strategy used), and outputs a set of prompts. These prompts are then fed into the assessment model $\mathbf{A}$ to generate an output $o \in \mathcal{O}$ (typically a JSON string), which is then parsed into an estimate for $E(c,p)$.

\subsection{Models}
\label{sec:llms}

For our assessment model $\mathbf{A}$, we consider the proprietary models GPT-3.5 \citep{brown2020language} and GPT-4 \citep{gpt4}, as well as the popular open source models Llama-2-70b-32k \citep{peng2023yarn} and Mixtral-8x7B \citep{jiang2024mixtral}. We run GPT-3.5 and GPT-4 via a secure Azure PHI-compliant instance. We run the open source models on a local PHI-compliant server with 4 Nvidia A100 GPUs using the vLLM \citep{vllm} and Outlines \citep{willard2023efficient} packages for inference and enforcing valid JSON output.

\subsection{Zero-Shot Criteria Assessment}
\label{sec:zero_shot}

We begin by evaluating the zero-shot capabilities of our assessment LLMs. We choose the zero-shot setting as it offers the simplest path towards real-world deployment -- it requires minimal engineering, no data labeling, and can instantly be adapted to any trial.

Given a set of notes $\mathcal{N}_p \subseteq \mathcal{N}$ for a patient $p$, we investigate four strategies for generating the prompt(s) that we feed into the assessment LLM: (1) \textit{All Criteria, All Notes (ACAN)} in which all notes in $\mathcal{N}_p$ are merged into a single prompt and the model generates all $C = |\mathcal{C}|$ eligibility assessments at once. (2) \textit{All Criteria, Individual Notes (ACIN)} in which the model is given $N = |\mathcal{N}_p|$ separate prompts (each containing one note), and for each note generates all $C$ eligibility assessments. (3) \textit{Individual Criteria, All Notes (ICAN)} in which all $N$ notes are merged into a single prompt, but the model only assesses one criterion at a time, and thus must be re-prompted with the same merged set of notes $C$ times. (4) \textit{Individual Criteria, Individual Notes (ICIN)} in which the model is given $N * C$ separate prompts each containing one note and one criterion. More precisely, we thus have:

\begin{align*}
    ACAN:~& \phi(\mathcal{N}_p^{(1)} \mathbin\Vert ... \mathbin\Vert \mathcal{N}_p^{(N)}, c_1  \mathbin\Vert ... \mathbin\Vert c_{C}) = t\\
    ACIN:~& \phi(\mathcal{N}_p^{(1)},..., \mathcal{N}_p^{(N)}, c_1 \mathbin\Vert ... \mathbin\Vert c_{C}) = \{ t_1, ..., t_N \}\\
    ICAN:~& \phi(\mathcal{N}_p^{(1)} \mathbin\Vert ... \mathbin\Vert \mathcal{N}_p^{(N)}, c_1, ..., c_{C}) = \{ t_1, ..., t_C \}\\
    ICIN:~& \phi(\mathcal{N}_p^{(1)},..., \mathcal{N}_p^{(N)}, c_1, ..., c_{C}) = \{ t_1, ..., t_{NC} \}\\
\end{align*}

Where $\mathbin\Vert$ represents concatenation. Please see Appendix Figures \ref{fig:prompt_template_all_criteria} and  \ref{fig:prompt_template_inclusion_criteria} for the exact prompt templates used. \textit{ACIN} and \textit{ICIN} generate a collection of assessments over $N$ different notes for a given patient. 
To aggregate patient-level criteria results over the collection of documents, we use the methodology described in Appendix Section \ref{sec:response_aggregation}

The output $o \in \mathcal{O}$ that we expect from the Assessment model $\mathbf{A}$ is a JSON string containing five elements:

\begin{enumerate}
    \item \textbf{Criterion:} The specific criterion being assessed.
    \item \textbf{Medications:} The names of all current medications and supplements that the patient is taking.
    \item \textbf{Rationale:} Step-by-step reasoning as to why the patient does or does not meet the criterion.
    \item \textbf{Decision:} Output ``MET" if the patient meets the criterion, or it can be inferred that they meet the criterion with common sense. Output ``NOT MET" if the patient does not, or it is impossible to assess given the provided information.
    \item \textbf{Confidence:} ``Low", ``Medium", or ``High" confidence in prediction.
\end{enumerate}

We include the \textbf{Medications} element to help the model recall specific drugs when assessing certain criteria, as prior literature demonstrated the ability of LLMs to perform this extraction \citep{agrawal2022large}. \textbf{Rationale} was included to provide interpretability for a human reviewer. We ablate this in Section \ref{sec:results_zero_shot} and find its inclusion has a slightly negative impact on accuracy. For the Llama-2 model only, we removed Medications and Confidence from the prompt as we were unable to get it to generate valid completions.

\subsection{Retrieval Pipeline}

We design a two-stage pipeline that leverages a small embedding model $\mathbf{M}$ to pre-filter the set of notes $n \in \mathcal{N}$ that get fed into the more expensive assessment model $\mathbf{A}$. This pipeline is depicted in Figure \ref{fig:figure_1}.

We consider two models for $\mathbf{M}$: BGE for its high performance on retrieval benchmarks \citep{bge_embedding}, and MiniLM for its speed \citep{reimers-2019-sentence-bert}. These models are small and can be run locally and cheaply. In a deployment scenario, these models would be run once over all clinical text within an EHR.

To initialize our pipeline, we loop through every patient and embed all of their notes using our embedding model $\mathbf{M}$. We also embed each criterion. Thus, we get a database of embeddings $\mathcal{D}$ such that $\mathcal{D} = \{ \mathbf{M}(n)~\forall n \in \mathcal{N} \} \cup \{ \mathbf{M}(c)~\forall c \in \mathcal{C} \}$.

During inference we have two stages -- \textbf{Pre-Filtering} and \textbf{Assessment}, which slightly differ depending on the prompting strategy used. The \textit{ICAN} and \textit{ICIN} strategies work as follows. In \textbf{Pre-Filtering}, we are given a patient $p$ and individual criterion $c$. We calculate the cosine similarity between $\mathbf{M}(n)$ and $\mathbf{M}(c)$ for all $n \in N_p$. We keep the top-$k$ ranking notes and discard the rest. We repeat this step for each criterion $c \in \mathcal{C}$. In \textbf{Assessment}, we construct our prompts with our subset of top-$k$ notes using our chosen prompting strategy $\phi(n_1,...,n_k, c_1,...,c_C)$. We then feed the prompt into the Assessment LLM to generate our output $o = A(\phi(n_k, c_1,...,c_C))$. The pipeline is similar for \textit{ACAN} and \textit{ACIN}, but instead of processing each individual criterion $c$ separately, we first concatenate them together to form a single criterion $c_1 \mathbin\Vert ... \mathbin\Vert c_C$ and only run the \textbf{Pre-Filtering} step once.

In practice, we work at the level of ``chunks" instead of ``notes," where a chunk is simply a substring taken from a larger note whose length fits within the context window of our embedding model $\mathbf{M}$. In our case, this was 512 tokens.

In addition to reducing the number of tokens input into the assessment model $\mathbf{A}$, this retrieval-based approach would allow our system to scale to real-world health systems with millions of notes \citep{jiang2023health}, an important efficiency consideration that was underexplored in prior work.

\section{Evaluation}

We evaluate the ability of LLMs to assess whether a patient meets or does not meet a set of inclusion criteria.

\subsection{Dataset}

We chose the 2018 n2c2 Clinical Trial Cohort Selection challenge dataset for evaluation as it is the largest and most realistic publicly available benchmark for clinical trial patient matching \citep{stubbs2019cohort}. Please see Appendix Section \ref{sec:dataset_comparison} for a more detailed comparison to other benchmarks. This dataset has two parts: patients and eligibility criteria.

\textbf{(1) Patients}. There are 288 unique patients (subdivided into two splits: 202 train and 86 test). Each patient has 2-5 de-identified clinical notes, with an average of 2,711 words each. The patients are all diabetic, and were originally sourced from a combined database of patients from Mass General and Brigham and Women's hospitals. Details on the cohort selection process are provided in \citep{stubbs2015annotating} and \citep{kumar2015creation}.

\textbf{(2) Eligibility criteria}. There are 13 inclusion criteria which reflect commonly used eligibility criteria \citep{stubbs2019cohort}. These criteria, as well as their definitions, are listed in Appendix Table \ref{tab:criteria_definition}. Note that there are no exclusion criteria. 

All of this information is provided as unstructured text. Thus, the only way to match patients is via natural language processing. Note that this benchmark simulates a \textit{synthetic} trial -- i.e. no actual clinical trial enrolled these patients. To obtain labels, two medical experts annotated each patient as to whether the 13 inclusion criteria were ``MET" or ``NOT MET". The Cohen's Kappa between the two annotators was 0.54 \citep{stubbs2019cohort}.

\subsection{Metrics} 

We evaluate our models on their precision, recall, and overall Macro/Micro-F1 scores on the binary classification tasks of deciding whether a patient met or did not meet each of the 13 inclusion criteria in the 2018 n2c2 benchmark. ``Overall" is defined by the n2c2 benchmark as the simple average of the F1 score calculated for the ``met" examples and F1 score for the ``not met" examples \citep{stubbs2019cohort, sota}. We use this definition for consistency when reporting Macro/Micro-F1 scores as well.

We also measure several efficiency metrics: (1) total number of tokens (both prompt and completion) used by the LLM; (2) total cost (in USD) of making predictions; (3) and total number of calls made to the LLM (as a proxy for time).

\subsection{Baselines}

In addition to LLMs, we consider two baselines. First, the best model from the original 2018 n2c2 competition, which achieved an Overall Micro-F1 of 0.91 \citep{sota}. We refer to this model as ``Prior SOTA." It utilized hand-crafted rules, negation detection, family history detection, and regular expressions \citep{sota}. We also consider predicting the class prevalence for each criterion, and refer to this as ``Baseline".

\subsection{Interpretability}

Unlike the rule-based Prior SOTA, our LLMs are able to output natural language rationales justifying their decisions. We recruit two clinicians to manually review the rationales generated by our best-performing model (GPT-4). Specifically, for each of the 13 criteria, we sample 20 patients for which GPT-4 made an incorrect prediction and 20 patients for which GPT-4 made a correct prediction. For criterion where GPT-4 made less than 20 incorrect predictions, we simply include all mispredictions. This gave us a total of 486 rationales.

\section{Results}

We evaluate our LLM-based systems on the 2018 n2c2 Cohort Selection Challenge dataset \citep{stubbs2019cohort}.

\subsection{Zero-Shot Criteria Assessment}
\label{sec:results_zero_shot}

For our initial zero-shot evaluation, we follow the \textit{ACIN} prompting strategy by feeding each note separately into the LLM and have it predict all criteria at once. All of the models we test are able to fit each patient's history into their context windows. We report our results in Table \ref{tab:zero_shot}. Despite not being tuned for trial matching or provided any in-context examples, GPT-4 beats the state-of-the-art by a margin of +6 Macro-F1 and +2 Micro-F1 points. We provide criterion-level confusion matrices for this model in Appendix Figure \ref{fig:confusion_matrices}. The open source models perform worse than GPT-3.5, and GPT-3.5 performs significantly worse than GPT-4  \citep{bubeck2023sparks}. This aligns with prior evaluations of LLMs on EHR-based tasks \citep{fleming2023medalign}, and suggests that prior trial matching work based on GPT-3.5 has significantly underestimated the capabilities of LLMs for this task \citep{trialgpt}. 

\begin{table*}[h!]
\centering
    \begin{tabular}{lrrrr}
    \toprule
    Model & Prec. & Rec. & \makecell{Overall\\Macro-F1} & \makecell{Overall\\Micro-F1} \\
    \midrule
    Baseline   & 0.69 & 0.78 & 0.43 & 0.76 \\
    Prior SOTA & 0.88    & 0.91    & 0.75 & 0.91 \\
    \hline
    Llama-2-70b & 0.82 & 0.41 & 0.46 & 0.67 \\
    Mixtral-8x7B & 0.72 & 0.83 & 0.64 & 0.79 \\
    \hline
    GPT-3.5 & 0.74 & 0.80 & 0.59 & 0.80 \\
    GPT-4   & \textbf{0.91} & \textbf{0.92} & \textbf{0.81} & \textbf{0.93} \\
    \bottomrule
    \end{tabular}
    \caption{\textbf{Zero-shot results on n2c2 2018 challenge}}
    \small
    \flushleft
    Zero-shot results on the 2018 n2c2 cohort selection challenge using the \textit{ACIN} prompt strategy. We use the 32k context length version of all models, with the exception of GPT-3.5 (which is limited to 16k tokens).
    \label{tab:zero_shot}
\end{table*}

While we have the model output a \textbf{Rationale} with each prediction for interpretability, we find that the inclusion of this field has a slightly negative impact on overall accuracy (results in Appendix Table \ref{tab:ablation_rationale}). As a brief initial comparison with few-shot prompting, we also evaluate the impact of including one example in the prompt. Surprisingly, we find in Appendix Table \ref{tab:few_shot_1} that our simple 1-shot prompting approach performs worse than zero-shot prompting. We hypothesize that this is due to the substantially higher number of tokens included in the 1-shot prompts (roughly 2.45x more for the \textit{ACIN} strategy), most of which are noise given the nature of clinical text. This makes it more difficult for GPT-4 to parse the relevant portions of the clinical note it is actually tasked with evaluating. Better prompting and fine-tuning strategies could certainly increase performance; however, given the focus of this paper is on the data/cost efficiency of LLMs, we leave such exploration to future work.

\subsection{Prompt Engineering}

The n2c2 benchmark does not come with a predefined set of prompts, but instead provides a brief definition for each criteria (see Appendix Table \ref{tab:criteria_definition}). As the low inter-rater agreement between clinicians suggests, these definitions did not fully capture how the dataset was labeled. Thus, to design our prompts we randomly sampled three patients who met and did not meet each criteria from the training dataset, and used this corpus to manually refine the definition of each criterion. We provide our ``improved" criteria in Appendix Table \ref{tab:criteria_improved}, and demonstrate in Table \ref{tab:prompt_eng} that increasing the specificity of criteria is an essential first step in using an LLM. In the real-world, this step could be avoided by directly using the detailed protocols provided by trial sponsors \citep{trialprotocol, trialprotocol2}.

\begin{table*}[t]
    \centering
    \begin{tabular}{llrrrr}
        \toprule
        Model & Prompt & Prec. & Rec. & \makecell{Overall\\Macro-F1} & \makecell{Overall\\Micro-F1} \\
        \midrule
        GPT-3.5 & Original & 0.76 & 0.48 & 0.57 & 0.69 \\
        & Improved & 0.76 & 0.82 & 0.63 & 0.82 \\~\\
        GPT-4 & Original & 0.89 & 0.74 & 0.75 & 0.85 \\
        & Improved & \textbf{0.91} & \textbf{0.92} & \textbf{0.81} & \textbf{0.93} \\
        \bottomrule
    \end{tabular}
    \caption{\textbf{Increasing prompt specificity improves performance}}
    \small
    \flushleft
    Increasing the specificity of the definitions for each criteria provided to the LLM can significantly improve the model's accuracy.
    \label{tab:prompt_eng}
\end{table*}

We ablate the 4 prompting strategies detailed in Section \ref{sec:zero_shot}. As shown in Table \ref{tab:prompt_strat}, the \textit{ACAN} strategy performs worst, while the rest of the strategies appear to be relatively equivalent in performance. When cost is considered, however, the \textit{ACIN} strategy becomes the clear winner -- \textbf{the \textit{ACIN} strategy achieves the best Micro-F1 at almost an order of magnitude lower cost, API calls, and tokens used} than the other prompting strategies. \textit{ACIN} passes each note through the LLM once and has it evaluate all criteria simultaneously, whereas the strategies relying on ``Individual Criteria" pass each note through the LLM 13 times (once for each criterion). Thus, careful prompting can result in meaningful cost savings with virtually no drop in accuracy.

\begin{table*}[t]
    \centering
    \scriptsize
    \begin{tabular}{lllrrrrrrr}
        \toprule
        \multicolumn{1}{l}{Model} & \multicolumn{2}{c}{Prompt Strategy} & \multicolumn{4}{c}{Performance} & \multicolumn{3}{c}{Efficiency} \\
         & Criteria & Notes & Prec. & Rec. & \makecell{Overall\\Macro-F1} & \makecell{Overall\\Micro-F1} & Cost & API Calls & Tokens \\
        \midrule
        GPT-3.5 & All & All & \underline{0.76} & 0.69 & 0.53 & 0.77 & \$1.24 & 86 & 0.64M \\
        & All & Individual & 0.74 & 0.80 & 0.59 & 0.80 & \$2.41 & 377  & 1.31M \\
        & Individual & All & \underline{0.76} & 0.82 & \underline{0.63} & \underline{0.82}  & \$13.19  & 1118 & 6.66M \\
        & Individual & Individual & 0.73 & \underline{0.88} & 0.60 & 0.81  &  \$17.61 & 4901  & 9.04M \\~\\
        GPT-4 & All & All & 0.89 & 0.87 & 0.80 & 0.90  & \$74.77 & 86  & 0.69M \\
        & All & Individual & 0.91 & \textbf{\underline{0.92}} & 0.81 & \textbf{\underline{0.93}}  & \$133.08 & 377  & 1.32M \\
        & Individual & All & \textbf{\underline{0.94}} & 0.86 & \textbf{\underline{0.85}} & 0.92  & \$780.99  & 1118 & 6.61M \\
        & Individual & Individual & 0.92 & 0.89 & 0.82 & 0.92  & \$1021.65  & 4901 & 8.93M \\
        \bottomrule
    \end{tabular}
    \caption{\textbf{Cost and data efficiency of prompting strategies}}
    \small
    \flushleft
    Considering one criterion/note at a time improves performance. ``Tokens" includes both prompt and completion tokens (i.e. inputs and outputs). ``API Calls" is the total number of times the LLM was queried. ``Cost" is based on OpenAI's pricing as of January 25, 2024 \citep{pricing}
    \label{tab:prompt_strat}
\end{table*}

\subsection{Retrieval Pipeline}

We test different cutoff values of $k$ for the number of chunks to retrieve, and measure its impact on GPT-4's performance under the \textit{ACAN}, \textit{ACIN}, and \textit{ICAN} prompting strategies. 

In Figure \ref{fig:retrieval}, we plot the performance of the model on the y-axis and the amount of tokens used on the x-axis. In these plots, the orange line is GPT-4, the blue line is GPT-3.5, the green line is the prior state-of-the-art, and the star is the non-retrieval-based implementation of each model. With our best embedding model, MiniLM, we are able to surpass the prior state-of-the-art on Macro-F1 using roughly one-third and one-half as many tokens as needed in the vanilla \textit{ICAN} and \textit{ACAN} strategies, respectively. While both methods approach the state-of-the-art on Micro-F1, neither retrieval-based method is able to fully achieve it without using the full note. There appears to be an unresolved gap in performance when relying on a retrieval-based system to curate patient information. Additionally, the cost benefits of the retrieval pipeline largely depend on the prompting strategy used -- as shown in Appendix Figure \ref{fig:retrieval_acin} since the \textit{ACIN} strategy feeds each chunk separately through the model, once $k$ exceeds the number of total notes in a patient's EHR, then the cost of using the retrieval pipeline ends up exceeding the cost of simply using full notes without retrieval. That is because each prompt contains a high number of non-note tokens (e.g. the instructions and criteria definitions), so splitting a note that could have been fit into one prompt into chunks spread across multiple separate prompts will increase the overall number of tokens used.

\definecolor{darkgreen}{rgb}{0.0, 0.5, 0.0}
\begin{figure*}[t]
    \includegraphics[width=1.0\textwidth]{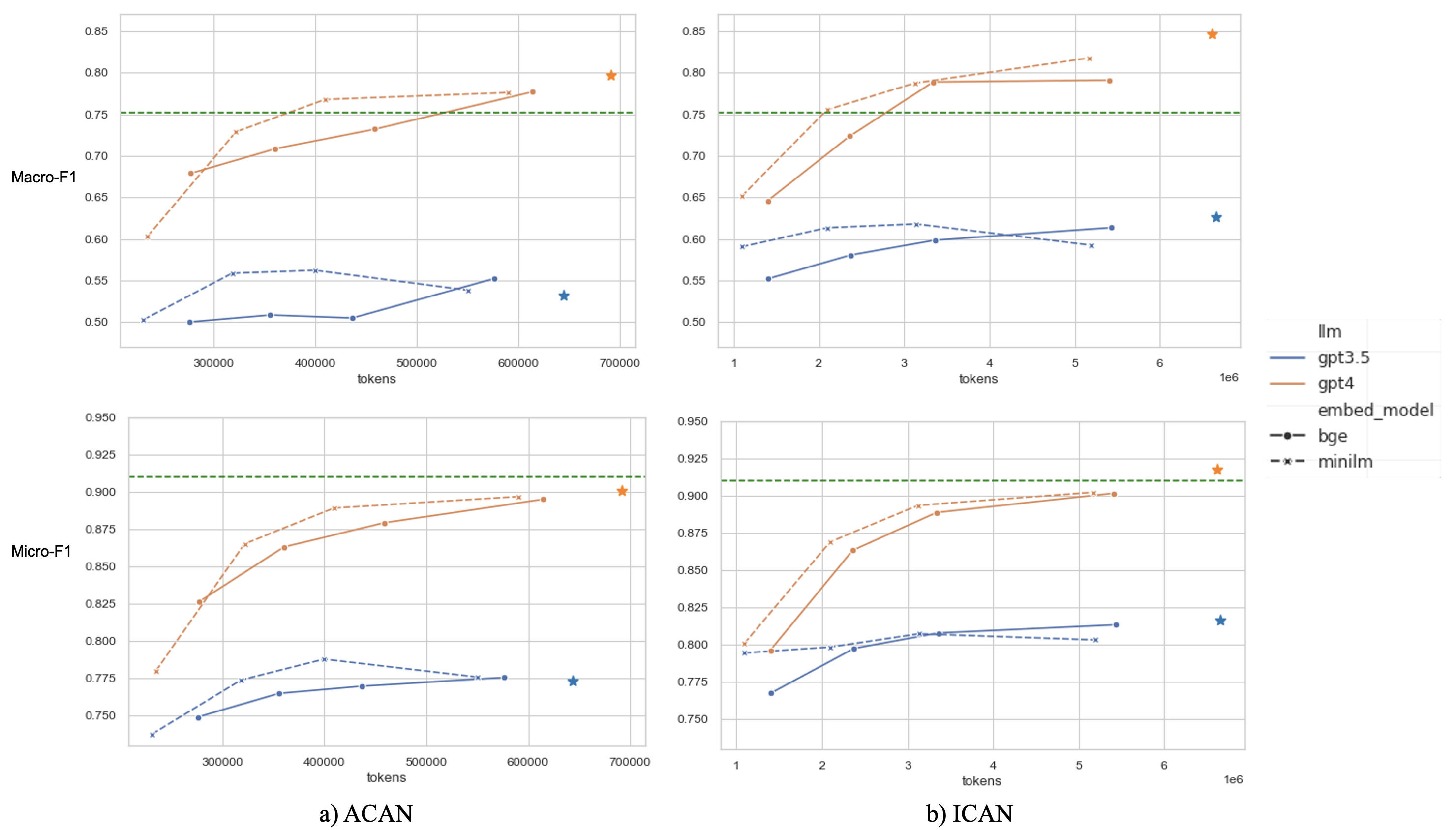}
    \centering
    \caption{\textbf{Two-stage retrieval pipeline performance}}
    \small
    \flushleft
    Model performance increases as the number ($k$) of retrieved chunks increases, but quickly plateaus with diminishing returns. We test $k \in \{1, 3, 5, 10 \}$. Each subfigure is a different prompting strategy. The y-axis is model performance (Macro/Micro-F1) and the x-axis is the total number of tokens processed by the model. \textcolor{orange}{Orange} is GPT-4, \textcolor{blue}{blue} is GPT-3.5, and the \textcolor{darkgreen}{green} line is the prior state-of-the-art. \textbf{Stars} represent each model's best performance when feeding in all notes. The MiniLM embedding model is the dashed line, while BGE is the solid line.    
    \label{fig:retrieval}
\end{figure*}

\subsection{Interpretability}

\begin{figure*}[t]
\floatconts
  {fig:subfigex}
  {\caption{\textbf{Clinician assessment of LLM-generated rationales}}}
  {%
    \subfigure[Correctness of rationales associated with inaccurate eligibility decisions by GPT-4]{
      \includegraphics[width=1\linewidth]{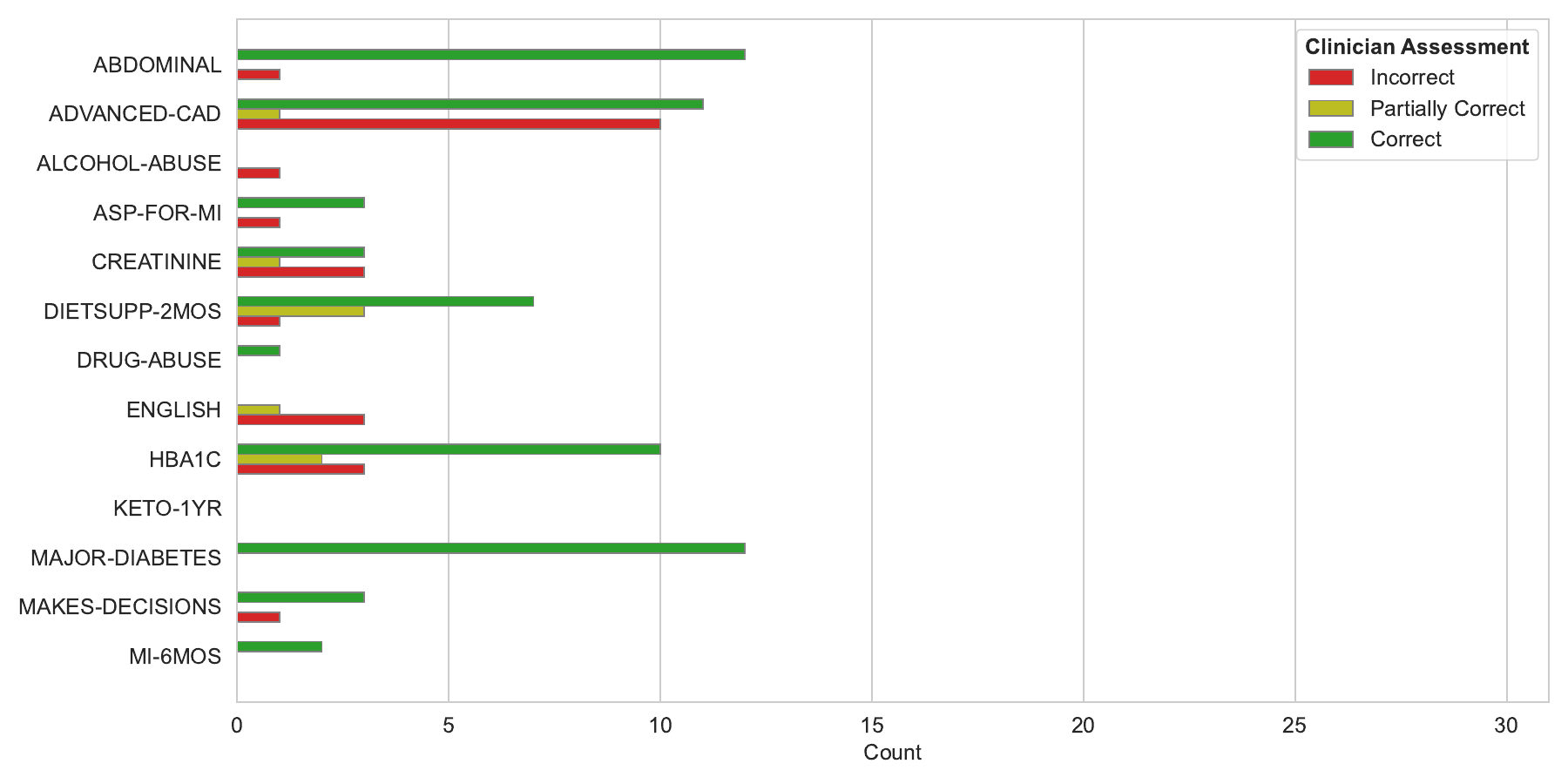}}
      \hfill
    \subfigure[Correctness of rationales associated with accurate eligibility decisions by GPT-4]{
      \includegraphics[width=1\linewidth]{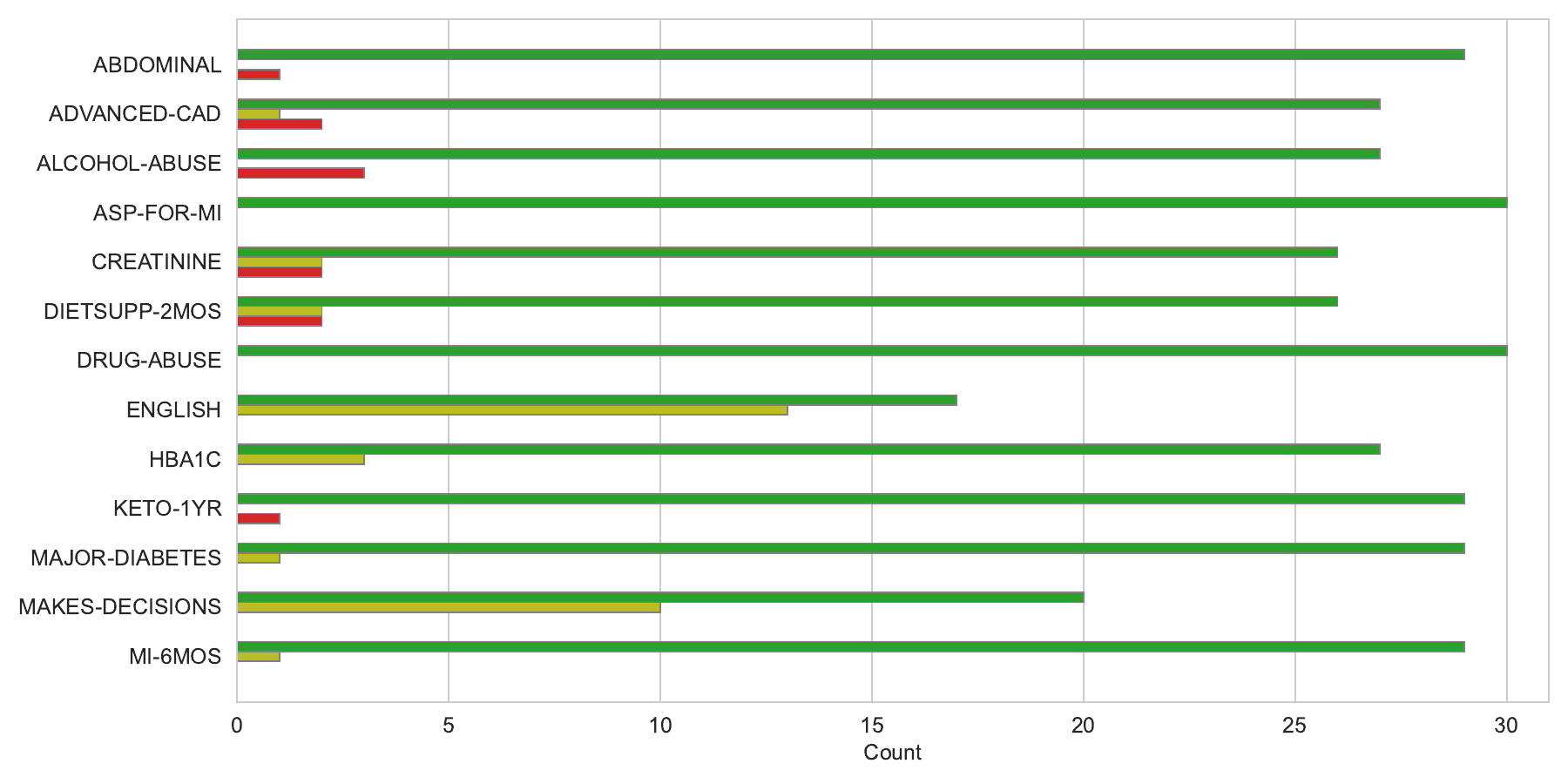}}%
  }
    \small
    \flushleft
    Clinician assessment of whether 486 randomly sampled rationales generated by GPT-4 were correct (\textcolor{green}{\textbf{green}}), partially correct (\textcolor{yellow}{\textbf{yellow}}), or incorrect (\textcolor{red}{\textbf{red}}), split by whether GPT-4's corresponding eligibility decision was evaluated as (a) incorrect or (b) correct according to the n2c2 benchmark. 
    \label{fig:clinican_rationale}
\end{figure*}

We sample 486 rationales generated by GPT-4 using the \textit{ICAN} prompting strategy, and have two clinicians evaluate their veracity. We use the \textit{ICAN} version of the system in order to cleanly disambiguate each individual criterion's prediction. Each rationale was evaluated on a 3-part scale: Correct, Partially Correct, and Incorrect, based on how accurately it aligned to the relevant patient's EHR. 

The results in Figure \ref{fig:clinican_rationale} show that GPT-4 is able to provide legitimate rationales for most its decisions. When GPT-4 makes a correct eligibility decision, 89\% of its rationales were judged as fully correct, 8\% as partially correct, and 3\% as incorrect. When GPT-4 made an incorrect eligibility decision, its rationales were split 67/8/25\% between correct/partially correct/incorrect. This indicates that either the ``ground truth" label of the n2c2 benchmark is incorrect, or that there is legitimate variation in clinical judgement as to whether the patient meets a criterion. This nuance is something that an LLM can capture through natural language in contrast to traditional rule-based and deep learning models. 

The most common errors noted by our clinician evaluators were pulling outdated medications and lab test values when mentioned across multiple notes (e.g. HBA1C values), inaccurately assessing the language of the patient when mentioned in a note, and over/under-weighting the severity of indicators of cardiovascular disease.

\section{Discussion}

In this work, we developed a zero-shot LLM-based pipeline to match patients to clinical trials. We achieved state-of-the-art performance on the 2018 n2c2 cohort selection challenge, achieving gains of +6 Macro-F1 and +2 Micro-F1 points over the prior best model without any feature engineering. Unlike prior methods, our system also provides interpretable rationale for its predictions, thereby allowing for human oversight and collaboration. Clinician evaluators disagreed with the LLM-generated rationales only 3\% of the time when the LLM made a correct eligibility decision, and 25\% of the time when it made an incorrect decision. Finally, we show that a preliminary retrieval step can preserve high performance while reducing token usage, and that thoughtful prompting can lead to large cost savings.

In contrast to prior work, we do no fine-tuning or updates to the underlying LLM's weights. Nor do we provide in-context few-shot examples. This illustrates the unique scalability advantages that LLMs such as GPT-4 offer to health systems, as they can be seamlessly repurposed for arbitrary trials so long as definitions for the criteria are known. Stanford, for example, has over 250 active clinical trials in oncology alone. Thus, even if a few-shot or fine-tuned system only required 5-10 hand-labeled patients per trial, deploying such an approach would require 1,000’s of labels, plus constant updating with each new trial. This is avoided with a zero-shot approach.

If our results generalize, then \textbf{LLMs represent a viable solution for clinical trial patient identification} given their accuracy, efficiency, and cost-effectiveness. Our GPT-4-based system is \textbf{accurate}, as shown in Table \ref{tab:zero_shot}, and is able to output coherent explanations for the majority of its decisions, as depicted in Appendix Figure \ref{fig:clinican_rationale}. It is \textbf{an order of magnitude faster} than today's clinical trials operate, as it was able to evaluate all 86 patients in roughly two hours versus the one patient per hour expected for Phase III trials \citep{penberthy2012effort}. Our pipeline is also \textbf{more cost-effective}, as Table \ref{tab:prompt_strat} shows that the cost for using GPT-4 with the \textit{ACIN} strategy to screen a single patient is roughly \$1.55, whereas prior estimates for a Phase III trial peg the cost per patient at \$34.75\footnote{See Table 3, \$$38193.70 / 1099 = \$34.75$ cost per patient screened.} \citep{penberthy2012effort}, a gap that will likely widen as models become more efficient.

Relying on LLMs poses risks as they have a propensity to "hallucinate", i.e. make up facts \citep{umapathi2023med}. Thus, we envision such an LLM-based trial matching system being deployed primarily as a "pre-screener" that automatically flags all patients in an EHR across all possible trials, thereby allowing clinical research coordinators to focus only on patients with strong enrollment potential. This minimizes the harm of misclassifying a patient by providing a natural means for human-in-the-loop validation.

\paragraph{Limitations} There are major limitations of our work. First, the size of our dataset is dwarfed by the scale of most health systems' EHRs, which underscores the need for using preliminary retrieval-based pipelines which can reduce the amount of tokens that an LLM must process to match each patient. Second, while the benchmark we use -- the 2018 n2c2 cohort challenge -- is the most realistic and largest public clinical trial matching dataset currently available, it is still a significant simplification of actual trial matching. It only covers inclusion criteria, only includes English, and its formatting might not generalize to other EHRs. Third, as shown in Table \ref{tab:prompt_eng}, the specificity of criteria definitions has a large impact on the Assessment LLM's ability to accurately assess patients. While this level of prompt engineering was still significantly easier than hand-engineering features with traditional NLP tools, and a real-world deployment would be able to leverage the much more detailed criteria definitions in a sponsor's trial protocol, these results do still raise a generalization concern. Fourth, although our LLM-based system achieved state-of-the-art results on this benchmark, further evaluations are required to asses the safety and failure modes of applying these models in production settings. Fifth, the use of proprietary models such as GPT-4 may not be feasible for health systems that do not have HIPAA-compliant Azure instances.  At a higher level, there are also open questions as to whether low enrollment in trials is a function of not being able to match patients (which this work addresses) or is instead primarily due to the stringency of the eligibility criteria themselves \citep{liu2021evaluating}. While beyond the scope of this work, we believe that improving the design of eligibility criteria presents another interesting research direction which could benefit from the application of more sophisticated models such as LLMs \citep{liu2021evaluating}.

\section{Conclusion} 

As our results demonstrate, LLMs have the potential to accelerate the matching of patients to clinical trials through their robust reasoning and natural language capabilities. Given the central importance of patient recruitment to drug discovery, we hope this work inspires further exploration of applying LLMs to clinical trial operations.

\acks{
This work was supported in part by the Clinical Excellence Research Center at Stanford Medicine and Technology and Digital Solutions at Stanford Healthcare. MW is supported by an HAI Graduate Fellowship. AL is supported by an ARC Institute Graduate Fellowship.
}

\bibliography{bib.bib}

\newpage
\appendix
\newpage
\setcounter{figure}{0}
\renewcommand{\thefigure}{S\arabic{figure}}

\setcounter{table}{0}
\renewcommand{\thetable}{S\arabic{table}}

\section{Compute Environment}
Experiments are performed in a local on-prem university compute environment using 24 Intel Xeon 2.70GHz CPU cores, 8 Nvidia V100 GPUs, 4 Nvidia A100 GPUs, and 1.48 TB of RAM. All compute environments supported HIPAA-compliant data protocols.

\section{Dataset Selection}
\label{sec:dataset_comparison}

The n2c2 2018 cohort selection benchmark is simplified from real-world patient matching, and curating more realistic datasets is a key opportunity for the field \citep{stubbs2019cohort}. However, after considering multiple datasets for this paper, we selected the n2c2 2018 dataset for four primary reasons: (1) it is the largest publicly available trial matching dataset available in terms of both patients and depth of medical history available for each patient; (2) it is the only benchmark that contains full clinical notes rather than short <10 sentence case reports, and is thus the most "realistic" for simulating the type of data present in a typical hospital's EHR; (3) by virtue of being publicly available, it enables apples-to-apples comparisons with prior work and encourages reproducible science; (4) it provides criteria-level eligibility labels and not just trial-level labels, which allows for more granular evaluation. We only considered public benchmarks in order to enable fair comparisons with prior work \citep{stubbs2019cohort}. The other three main datasets we've seen evaluated in the literature \citep{trialgpt, trialllama} are: TREC 2021 \citep{roberts2021overview}, TREC 2022 \citep{roberts2021overview}, and SIGIR 2016 \citep{koopman2016test}. Compared to the n2c2 2018 benchmark, however, these three datasets all have fewer patients (75, 50, and 60 respectively) and significantly less textual information associated with each patient (the patients are presented via "case reports," which are dense 5-10 sentence summaries of the patient's entire relevant medical history written in plain English. This makes these datasets a poor fit for the task of clinical trial patient matching, as case reports are significantly different than the clinical text that would actually be encountered by a clinical trial matching system if it were deployed in a hospital setting. In contrast, the n2c2 dataset sources its text directly from deidentified clinical notes and thus offers a much more realistic distribution of text. Additionally, the brevity of case reports defeats the motivation for doing efficient retrieval and prompting (which was the focus of this work) -- if you can easily fit everything in the context window of an LLM, then there is no need for the methods of this work. For example, the entire TREC 2021 dataset contains only 15,066 tokens of text, which means two entire copies of this dataset could fit into a single prompt of GPT-4, with room to spare \citep{roberts2021overview}.

\section{Response Aggregation}
\label{sec:response_aggregation}

When using the \textit{ACIN} and \textit{ICIN} prompting strategies, we get one eligibility prediction per criterion per note. This means a patient will have multiple eligibility decisions per criterion, and we therefore need a way to aggregate these decisions into one single decision per patient. To perform this aggregation, we use the following methodology on a criterion-by-criterion basis:

\begin{enumerate}
\item Criteria concerning the existence of a medical event across a patient's entire medical history (e.g. having had an intra-abdominal surgery) are considered "MET" if at least one note results in a "MET" assessment by the LLM.
\item Criteria which are time-dependent (e.g. having taken a dietary supplement in the past 2 months) use the eligibility decision of the most recent note within the relevant time window.
\end{enumerate}

The specific strategy used for each criterion is shown below.

\begin{table}[htbp]
    \centering
    \begin{tabular}{ll}
        \toprule
        \textbf{Criteria} & \textbf{Aggregation Method} \\
        \midrule
        ABDOMINAL & max \\
        ADVANCED-CAD & max \\
        ASP-FOR-MI & max \\
        CREATININE & max \\
        DRUG-ABUSE & max \\
        ALCOHOL-ABUSE & max \\
        HBA1C & max \\
        MAJOR-DIABETES & max \\
        MAKES-DECISIONS & max\\
        ENGLISH & min \\
        DIETSUPP-2MOS & most recent (2 months) \\
        KETO-1YR & most recent (12 months) \\
        MI-6MOS & most recent (6 months) \\
        \bottomrule
    \end{tabular}
    \label{tab:criteria_agg}
    \caption{\textbf{Criterion aggregation strategies}}
    \small
    \flushleft
    Criterion aggregation methods for each inclusion criterion in the nc2c 2018 dataset. 
\end{table}

\begin{figure*}[h]
    \centering
    \includegraphics[width=0.77\textwidth]{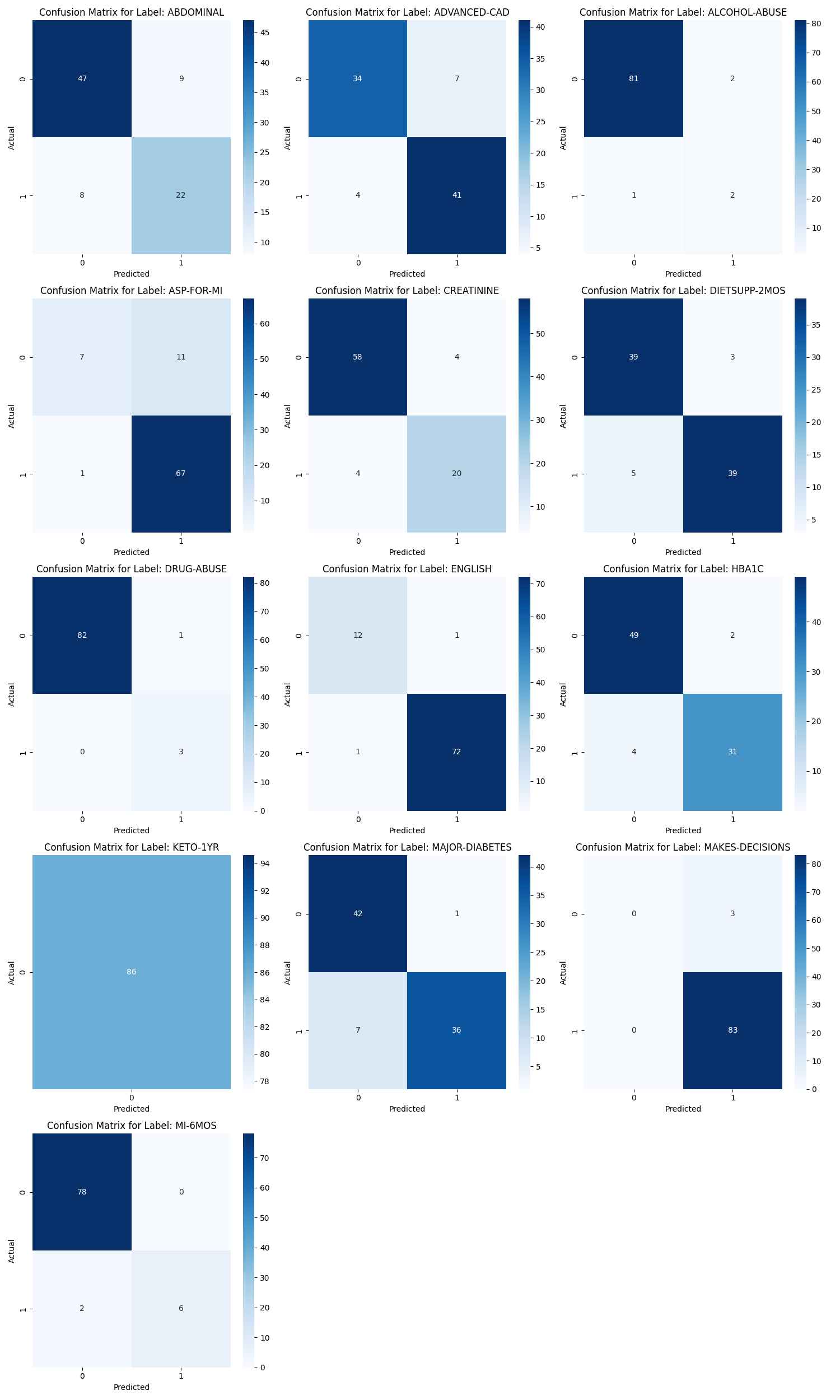}
    \caption{\textbf{Criterion-level confusion matrices}}
    \small
    \flushleft
    Confusion matrices for all criteria for our best performing model, GPT-4 using the \textit{ACIN} prompt.
    \label{fig:confusion_matrices}
\end{figure*}

\begin{table*}[h]
    \small
    \centering
    \begin{tabular}{ccccccc}
        \hline
        Criteria  &  True Value  & Prec. & Rec. & F1 & Support \\
        \hline
        ABDOMINAL    & MET & 0.7097  & 0.7333 & 0.7213     & 30      \\
        ABDOMINAL    & NOT MET & 0.8545  & 0.8393 & 0.8468     & 56      \\
        ADVANCED-CAD & MET &   0.8542 & 0.9111  &  0.8817 & 45      \\
        ADVANCED-CAD &  NOT MET &  0.8947  & 0.8293 &  0.8608       & 41      \\
        ALCOHOL-ABUSE & MET &   0.5000 &   0.6667   &   0.5714    & 3       \\
        ALCOHOL-ABUSE & NOT MET &  0.9878 &  0.9759 &  0.9818      & 83      \\
        ASP-FOR-MI & MET &   0.8590   & 0.9853   &  0.9178     & 68      \\
        ASP-FOR-MI & NOT MET &     0.8750 &  0.3889 &  0.5385     & 18      \\
        CREATININE & MET &    0.8333  &  0.8333 &  0.8333    & 24      \\
        CREATININE & NOT MET &   0.9355 &  0.9355 &  0.9355     & 62      \\
        DIETSUPP-2MOS & MET &   0.9286  &  0.8864  &    0.9070      & 44      \\
        DIETSUPP-2MOS & NOT MET &   0.8864 &  0.9286 &  0.9070   & 42      \\
        DRUG-ABUSE  & MET &   0.7500   & 1.0000   &  0.8571     & 3       \\
        DRUG-ABUSE  &  NOT MET &    1.0000  &  0.9880 &  0.9939    & 83      \\
        ENGLISH & MET &   0.9863  & 0.9863   &   0.9863     & 73      \\
        ENGLISH &  NOT MET &     0.9231 &  0.9231  &  0.9231   & 13      \\
        HBA1C & MET &     0.9394 &   0.8857   &  0.9118    & 35      \\
        HBA1C &  NOT MET &    0.9245 &  0.9608 &  0.9423    & 51      \\
        KETO-1YR & MET &    0.0000  &  0.0000  &  0.0000    & 0       \\
        KETO-1YR & NOT MET &   1.0000  & 1.0000  & 1.0000     & 86      \\
        MAJOR-DIABETES & MET &    0.9730  &  0.8372  &   0.9000    & 43      \\
        MAJOR-DIABETES & NOT MET &   0.8571 &  0.9767  & 0.9130     & 43      \\
        MAKES-DECISIONS & MET &    0.9651  &  1.0000 &   0.9822    & 83      \\
        MAKES-DECISIONS & NOT MET &    0.0000  & 0.0000  & 0.0000     & 3       \\
        MI-6MOS & MET &    1.0000 &   0.7500  &  0.8571   & 8       \\
        MI-6MOS & NOT MET &    0.9750  & 1.0000  & 0.9873     & 78      \\
        \hline
    \end{tabular}
    \caption{\textbf{Criterion-level accuracies}}
    \small
    \flushleft
    Criterion-level breakdown for our best model (GPT-4 using the \textit{ACIN} prompt)
    \label{tab:best_model_criteria_performance}
\end{table*}

\begin{figure*}[h]
    \centering
    \includegraphics[width=0.9\textwidth]{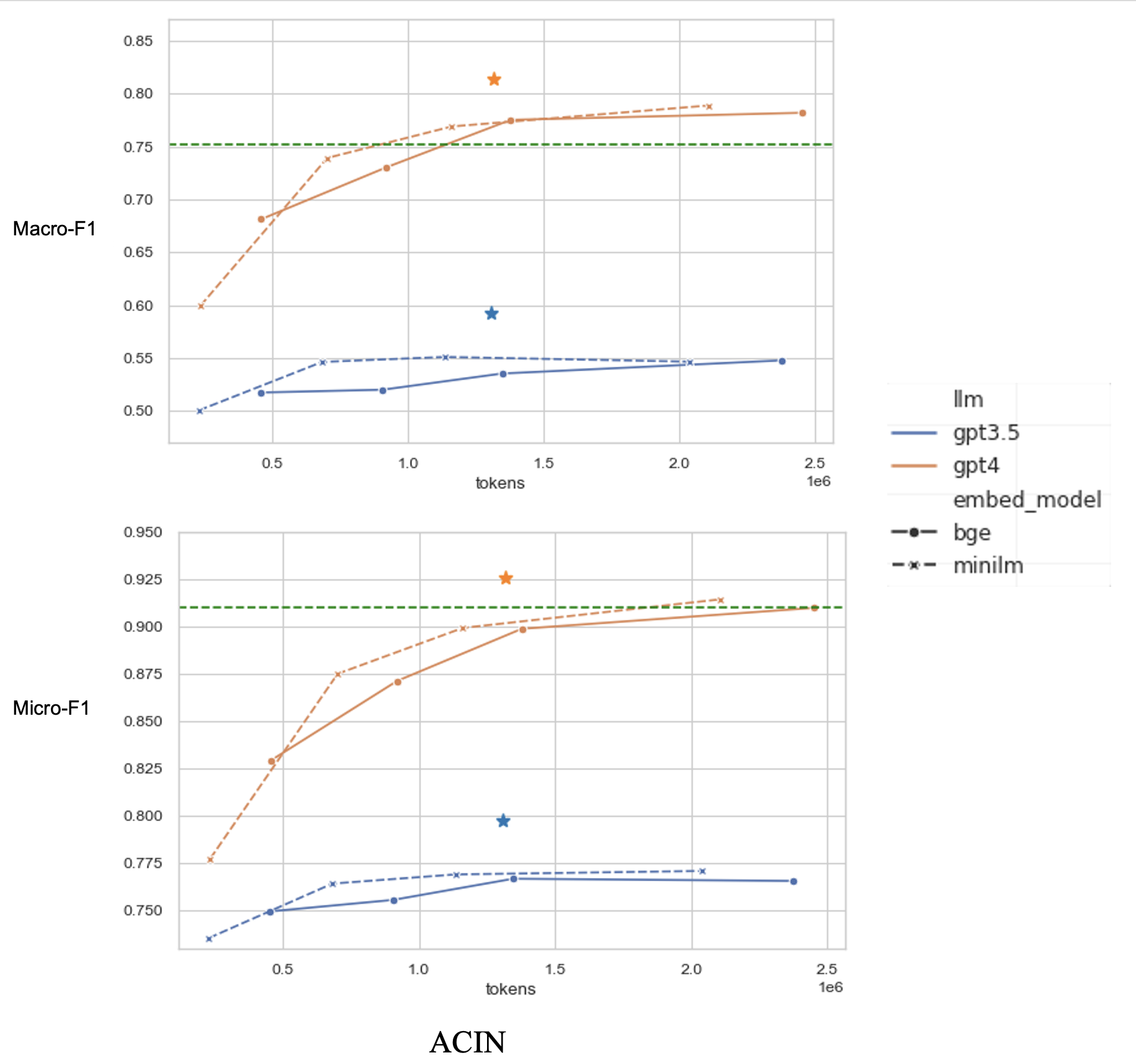}
    \caption{\textbf{Two-Stage Retrieval with $ACIN$ strategy}}
    \small
    \flushleft
    We test $k \in \{1, 3, 5, 10 \}$ for the $ACIN$ prompting strategy. Doing retrieval with $ACIN$ can actually be less cost-efficient to achieve the same level of performance. That is because the \textit{ACIN} strategy feeds each chunk separately through the model. Thus, once $k$ exceeds the number of total notes in a patient's EHR, the cost of using the retrieval pipeline ends up exceeding the cost of simply using full notes without retrieval.
    \label{fig:retrieval_acin}
\end{figure*}

\begin{table*}[t]
    \centering
    \begin{tabular}{lllrrrr}
        \toprule
        \multicolumn{1}{l}{Model} & \multicolumn{2}{c}{Prompt Strategy} & \multicolumn{4}{c}{Change in Performance} \\
         & Criteria & Notes & Prec. & Rec. & \makecell{Overall\\Macro-F1} & \makecell{Overall\\Micro-F1} \\
        \midrule
        GPT-3.5 & All & All & 0.04 & 0.03 & 0.05 & 0.03 \\
        GPT-3.5 & All & Individual & 0.07 & 0.01 & 0.05 & 0.04\\
        GPT-3.5 & Individual & All & 0.05 & 0.03 & 0.05 & 0.03 \\
        GPT-3.5 & Individual & Individual & 0.00 & -0.01 & 0.01 & 0.01\\~\\
        GPT-4 & All & All & -0.01 & 0.03 & 0.00 & 0.01 \\
        GPT-4 & All & Individual & -0.04 & 0.01 & -0.02 & -0.02 \\
        GPT-4 & Individual & All & -0.02 & 0.02 & -0.01 & -0.01 \\
        GPT-4 & Individual & Individual & 0.00 & 0.02 & 0.01 & -0.01 \\
        \bottomrule
    \end{tabular}
    \caption{\textbf{Impact of \textit{Rationale} field}}
    \small
    \flushleft
    Change in performance after removing the \textbf{Rationale} field from the prompt (compared to scores in Table \ref{tab:prompt_strat})
    \label{tab:ablation_rationale}
\end{table*}

\begin{table*}[t]
    \centering
    \begin{tabular}{lllrrrr}
        \toprule
        \multicolumn{1}{l}{Model} & \multicolumn{2}{c}{Prompt Strategy} & \multicolumn{4}{c}{Change in Performance} \\
         & Criteria & Notes & Prec. & Rec. & \makecell{Overall\\Macro-F1} & \makecell{Overall\\Micro-F1} \\
        \midrule
        GPT-4 & All & All & -0.01 & -0.02 & -0.04 & -0.02 \\
        GPT-4 & All & Individual & -0.07 & 0.01 & -0.05 & -0.03 \\
        GPT-4 & Individual & All & -0.01 & -0.03 & -0.06 & -0.02  \\
        GPT-4 & Individual & Individual & -0.01 & 0.02 & -0.01 & 0.00 \\
        \bottomrule
    \end{tabular}
    \caption{\textbf{Impact of 1-shot example}}
    \small
    \flushleft
    Change in performance after adding a 1-shot example to the prompt (compared to Table \ref{tab:prompt_strat})
    \label{tab:few_shot_1}
\end{table*}

\renewcommand{\arraystretch}{2}
\begin{table*}[ht!]
  \small
  \centering 
  \begin{tabularx}{1\textwidth}{|p{4cm}|X|p{2cm}|}
    \hline
    \textbf{Criteria} & \textbf{Definition} & \textbf{Prevalence}\\
    \hline
    ABDOMINAL & History of intra-abdominal surgery, small or large intestine resection, or small bowel obstruction & 0.35 \\
    ADVANCED-CAD & Advanced cardiovascular disease (CAD). For the purposes of this annotation, we define “advanced” as having 2 or more of the following: • Taking 2 or more medications to treat CAD • History of myocardial infarction (MI) • Currently experiencing angina • Ischemia, past or present & 0.52 \\
    ALCOHOL-ABUSE & Current alcohol use over weekly recommended limits & 0.03 \\
    ASP-FOR-MI & Use of aspirin to prevent MI & 0.79 \\
    CREATININE & Serum creatinine $>$ upper limit of normal & 0.28 \\
    DIETSUPP-2MOS & Taken a dietary supplement (excluding vitamin D) in the past 2 months & 0.51 \\
    DRUG-ABUSE & Drug abuse, current or past & 0.03 \\
    ENGLISH & Patient must speak English & 0.85 \\
    HBA1C & Any hemoglobin A1c (HbA1c) value between 6.5\% and 9.5\% & 0.41 \\
    KETO-1YR & Diagnosis of ketoacidosis in the past year & 0.00 \\
    MAJOR-DIABETES & Major diabetes-related complication. For the purposes of this annotation, we define ``major complication" (as opposed to “minor complication”) as any of the following that are a result of (or strongly correlated with) uncontrolled diabetes: • Amputation • Kidney damage • Skin conditions • Retinopathy • nephropathy • neuropathy & 0.50 \\
    MAKES-DECISIONS &  Patient must make their own medical decisions & 0.97 \\
    MI-6MOS & MI in the past 6 months & 0.09 \\
    \hline
  \end{tabularx}
  \caption{\textbf{2018 n2c2 eligibility criteria description}}
  \small
  \flushleft
  Eligibility criteria in the 2018 n2c2 cohort selection dataset, along with the percentage of patients who meet each criterion in the test set. Definitions taken verbatim from \citep{stubbs2019cohort}.
  \label{tab:criteria_definition} 
\end{table*}

\renewcommand{\arraystretch}{2}
\begin{table*}[ht!]
  \scriptsize
  \centering 
  \begin{tabularx}{1\textwidth}{|p{4cm}|X|p{3cm}|}
    \hline
    \textbf{Criteria} & \textbf{Definition}\\
    \hline
    ABDOMINAL & History of intra-abdominal surgery. This could include any form of intra-abdominal surgery, including but not limited to small/large intestine resection or small bowel obstruction\\
    ADVANCED-CAD & Advanced cardiovascular disease (CAD). For the purposes of this annotation, we define “advanced” as having 2 or more of the following: (a) Taking 2 or more medications to treat CAD (b) History of myocardial infarction (MI) (c) Currently experiencing angina (d) Ischemia, past or present. The patient must have at least 2 of these categories (a,b,c,d) to meet this criterion, otherwise the patient does not meet this criterion. For ADVANCED-CAD, be strict in your evaluation of the patient -- if they just have cardiovascular disease, then they do not meet this criterion.\\
    ALCOHOL-ABUSE & Current alcohol use over weekly recommended limits\\
    ASP-FOR-MI & Use of aspirin for preventing myocardial infarction (MI).\\
    CREATININE & Serum creatinine level above the upper normal limit\\
    DIETSUPP-2MOS & Consumption of a dietary supplement (excluding vitamin D) in the past 2 months. To assess this criterion, go through the list of medications\_and\_supplements taken from the note. If a substance could potentially be used as a dietary supplement (i.e. it is commonly used as a dietary supplement, even if it is not explicitly stated as being used as a dietary supplement), then the patient meets this criterion. Be lenient and broad in what is considered a dietary supplement. For example, a 'multivitamin' and 'calcium carbonate' should always be considered a dietary supplement if they are included in this list.\\
    DRUG-ABUSE & Current or past history of drug abuse\\
    ENGLISH & Patient speaks English. Assume that the patient speaks English, unless otherwise explicitly noted. If the patient\'s language is not mentioned in the note, then assume they speak English and thus meet this criteria.\\
    HBA1C & Any hemoglobin A1c (HbA1c) value between 6.5\% and 9.5\%\\
    KETO-1YR & Diagnosis of ketoacidosis within the past year\\
    MAJOR-DIABETES & Major diabetes-related complication. Examples of “major complication” (as opposed to “minor complication”) include, but are not limited to, any of the following that are a result of (or strongly correlated with) uncontrolled diabetes: • Amputation • Kidney damage • Skin conditions • Retinopathy • nephropathy • neuropathy. Additionally, if multiple conditions together imply a severe case of diabetes, then count that as a major complication.\\
    MAKES-DECISIONS & Patient must make their own medical decisions. Assume that the patient makes their own medical decisions, unless otherwise explicitly noted. There is no information provided about the patient\'s ability to make their own medical decisions, then assume they do make their own decisions and therefore meet this criteria\\
    MI-6MOS & Myocardial infarction (MI) within the past 6 months\\
    \hline
  \end{tabularx}
  \caption{\textbf{Redefined criteria definitions}}
  \small
  \flushleft
  ``Improved" criteria definitions used in our prompts.
  \label{tab:criteria_improved} 
\end{table*}

\begin{figure*}[h]
    \centering
    \includegraphics[width=0.9\textwidth]{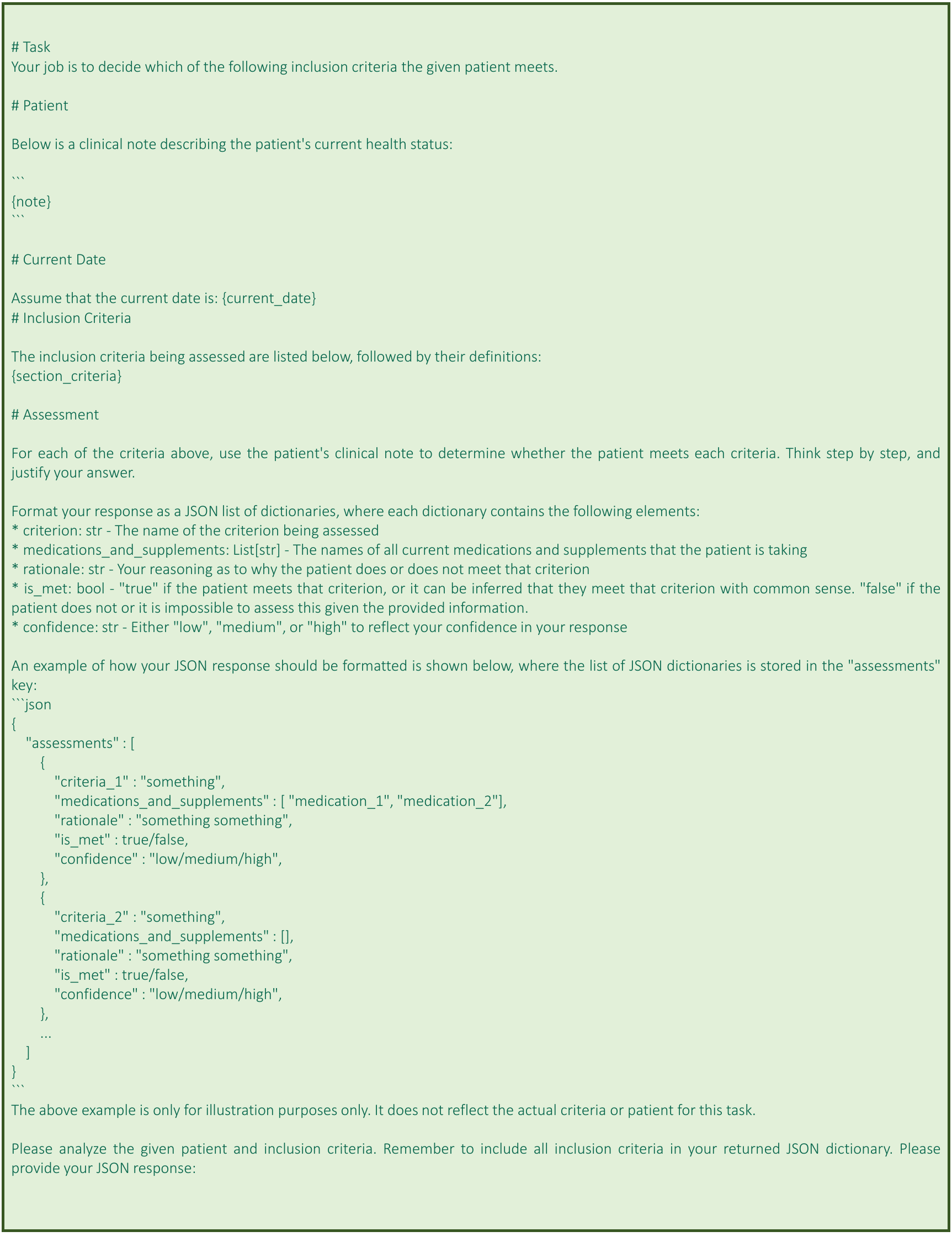}
    \caption{\textbf{Prompt used for All Criteria setting}}
    \small
    \flushleft
    The prompt template we use for the \textbf{All Criteria} setting, i.e. the \textit{ACIN} and \textit{ACAN} strategies. The only variables injected into this prompt are \textbf{note}, \textbf{section\_criteria}, and \textbf{current\_date}. Respectively, these contain the text from the patient's EHR, the definitions of the inclusion criteria being evaluated, and the date of the last note in the patient's EHR.
    \label{fig:prompt_template_all_criteria}
\end{figure*}

\begin{figure*}[h]
    \includegraphics[width=0.9\textwidth]{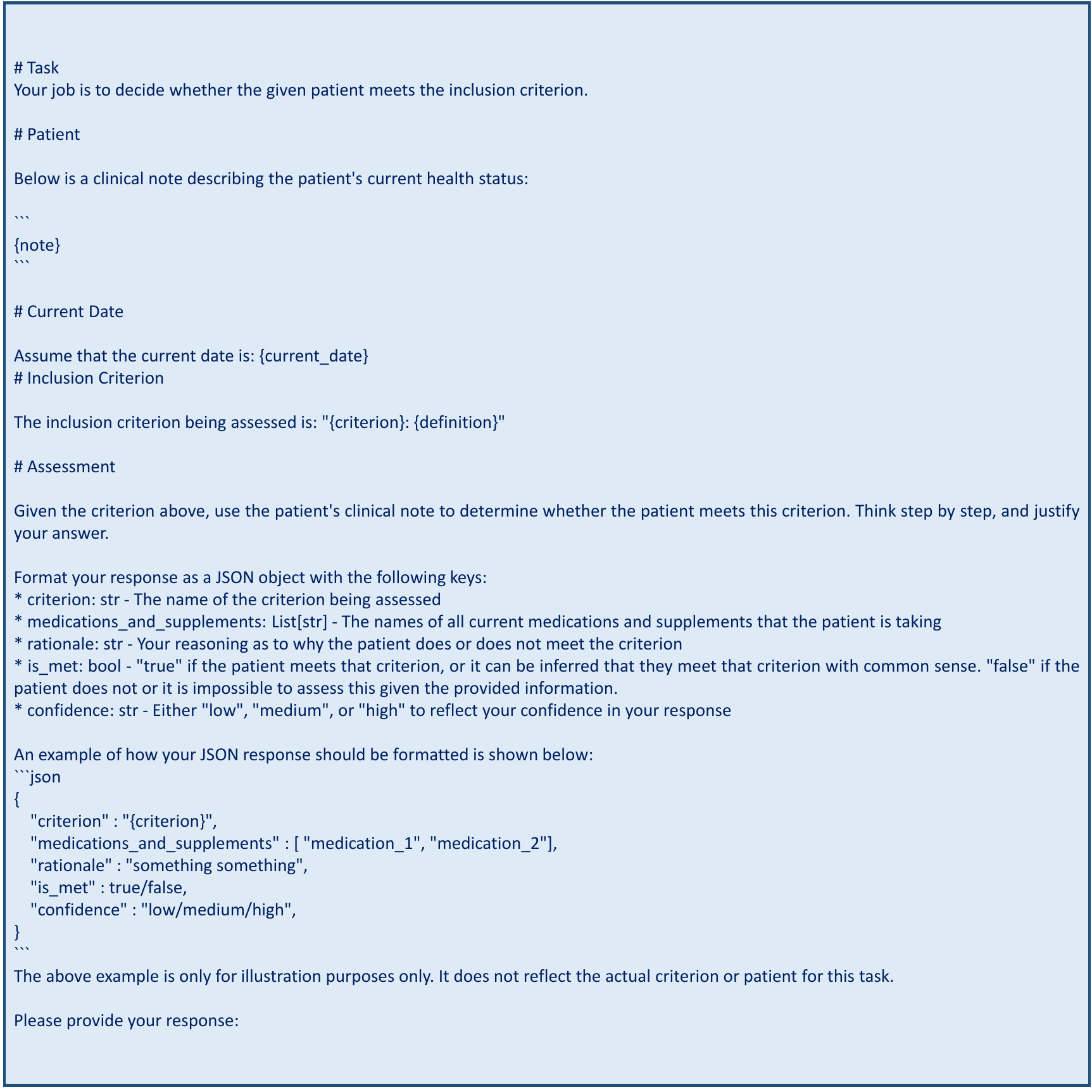}
    \caption{\textbf{Prompt used for Individual Criteria  setting}}
    \small
    \flushleft
    The prompt template we use for the \textbf{Individual Criteria} setting, i.e. the \textit{ICIN} and \textit{ICAN} strategies. The only variables injected into this prompt are \textbf{note}, \textbf{section\_criteria}, and \textbf{current\_date}. Respectively, these contain the text from the patient's EHR, the definitions of the inclusion criteria being evaluated, and the date of the last note in the patient's EHR.
    \label{fig:prompt_template_inclusion_criteria}
\end{figure*}

\end{document}